\documentclass[11pt,a4paper]{article}
\usepackage[hyperref]{naaclhlt2019}
\usepackage{times}
\usepackage{latexsym}
\usepackage{url}
\usepackage{titlesec}

\titleformat{\paragraph}
{\normalfont\normalsize\bfseries}{\theparagraph}{1em}{}
\titlespacing*{\paragraph}
{0pt}{3.25ex plus 1ex minus .2ex}{1.5ex plus .2ex}

\aclfinalcopy 
\def\aclpaperid{10} 

\title{CODAH: An Adversarially-Authored Question Answering Dataset for Common Sense}

\author{Michael Chen \quad Mike D'Arcy \quad Alisa Liu \quad Jared Fernandez \quad Doug Downey \\ Department of Computer Science \\ Northwestern University \\ Evanston, IL 60208 \\ 
{\small \texttt{\{y-chen, m.m.darcy, alisa, jared.fern\}@u.northwestern.edu}}\\  {\texttt{\small ddowney@eecs.northwestern.edu}}}

\date{June 1, 2019}

\begin{document}
\maketitle
\begin{abstract}
    Commonsense reasoning is a critical AI capability, but it is difficult to construct challenging datasets that test common sense. Recent neural question answering systems, based on large pre-trained models of language, have already achieved near-human-level performance on commonsense knowledge benchmarks. These systems do not possess human-level common sense, but are able to exploit limitations of the datasets to achieve human-level scores.
    
    We introduce the CODAH dataset, an adversarially-constructed evaluation dataset for testing common sense. CODAH forms a challenging extension to the recently-proposed SWAG dataset, which tests commonsense knowledge using sentence-completion questions that describe situations observed in video. To produce a more difficult dataset, we introduce a novel procedure for question acquisition in which workers author questions designed to target weaknesses of state-of-the-art neural question answering systems. Workers are rewarded for submissions that models fail to answer correctly both before {\em and} after fine-tuning (in cross-validation). We create 2.8k questions via this procedure and evaluate the performance of multiple state-of-the-art question answering systems on our dataset. We observe a significant gap between human performance, which is 95.3\%, and the performance of the best baseline accuracy of 67.5\% by the BERT-Large model.
\end{abstract}

\section{Introduction}
    Enabling commonsense reasoning in machines is a longstanding challenge in AI.  The rise of data-driven methods has led to interest in developing large datasets for commonsense reasoning over text. 

    The Situations With Adversarial Generations (SWAG) dataset \cite{zellers2018swag} introduced a large-scale benchmark for commonsense question answering in the form of multiple choice sentence completion questions describing situations as observed in video.  However, while SWAG was constructed to be resistant to certain baseline algorithms, powerful subsequent methods were able to perform very well on the dataset.  In particular, the development of the transformer architecture \cite{vaswani2017attention} has led to powerful pre-trained language model representations, including the OpenAI Transformer Language Model \cite{radford2018improving} and the Bidirectional Encoder Representations from Transformers (BERT) model \cite{devlin2018bert}. BERT achieved new state-of-the-art performance on SWAG that exceeded even that of a human expert.  However, BERT does not possess human-level common sense in general, as our experiments demonstrate.  It is instead able to exploit regularities in the SWAG dataset to score high. This motivates the construction of additional datasets that pose new challenges, and serve as more reliable benchmarks for commonsense reasoning systems.
    
    In this work, we introduce the \textbf{CO}mmonsense \textbf{D}ataset \textbf{A}dversarially-authored by \textbf{H}umans (\textbf{CODAH}) for commonsense question answering in the style of SWAG multiple choice sentence completion. We propose a novel method for question generation, in which human annotators are educated on the workings of a state-of-the-art question answering model, and are asked to submit questions that adversarially target the weaknesses. Annotators are rewarded for submissions in which the model fails to identify the correct sentence completion both before {\em and} after fine-tuning on a sample of the submitted questions, encouraging the creation of questions that are not easily learnable.
    
    We experimentally demonstrate that CODAH's generation procedure produces a dataset with a large gap between system performance and human expert accuracy, even when using state-of-the-art pre-trained language models with and without fine-tuning on the large SWAG dataset.  Using a model initially fine-tuned on SWAG, we find that the OpenAI GPT-1 and BERT neural question answering models yield 65.5\% and 69.5\% accuracy, respectively, on the CODAH dataset in cross-validation.  Thus, cross-validating on CODAH can form a challenging additional evaluation for SWAG-style commonsense QA systems.
    Human evaluators achieve 95.3\% accuracy, which is substantially higher than the 85.0\% \cite{zellers2018swag} and 87.7\% \cite{ghaeini2018dr} human performance on the SWAG and SNLI natural language inference tasks. The high human performance suggests that answers to the CODAH questions are in fact commonsense knowledge. Finally, we also analyze differences in performance across questions that target different types of commonsense reasoning, including quantitative, negation, and object reference, showing consistency in performance for BERT and GPT on the proposed categories.

\section{Related Work}
    Prior work in question answering has largely focused on the development of reading comprehension-based question answering and resulted in the creation of several large datasets for factoid extraction such as SQuAD \cite{rajpurkar2016squad, P18-2124} and the Google Natural Questions datasets  \cite{47761}. In these tasks, extraction of correct answers from the provided context requires little external world knowledge, understanding of intents, or other commonsense knowledge. 
    
    Earlier work has established multiple benchmarks for natural language inference and linguistic entailment with the release SNLI \cite{bowman2015large} and MultiNLI datasets \cite{williams2018broad}. In these tasks, systems must identify whether a hypothesis agrees with or contradicts a provided premise. In these datasets, determining entailment solely relies upon the provided premise and does not require a question answering system to utilize external knowledge. More recently, the SWAG dataset \cite{zellers2018swag} directly targets natural language inference that leverages commonsense knowledge. SWAG multiple choice completion questions are constructed using a video caption as the ground truth with incorrect counterfactuals created using adversarially-filtered generations from an LSTM language model.  State-of-the-art models for natural language inference have rapidly improved and approach human performance, which leaves little room for continued improvement on current benchmarks. 
    
    Generation of adversarial examples has also been used to increase the robustness of NLP systems as part of the \textit{Build it, Break It, The Language Edition} Workshop \cite{ettinger2017towards}. In this workshop, builders designed systems for Sentiment Analysis and Question Answering Driven Semantic Role Labeling tasks and were evaluated on the accuracy of their models on adversarial test cases designed by breakers. Whereas \textit{Build It Break It} adversarial generation required submissions to match the format of a starter dataset and offered limited adversarial access to the target NLP systems, the CODAH construction procedure allows for entirely new questions and provide adversaries with a target model throughout the submission process, allowing workers to experiment.

\section{The CODAH Dataset}
    Our dataset contains multiple choice sentence completion questions in the format of the SWAG dataset. Examples of the questions are shown in Table \ref{labels-table}.  Each question consists of a prompt sentence, the subject of the subsequent sentence, and four candidate completions, such that exactly one candidate completion is consistent with common sense. This task definition allows for easy evaluation by many state-of-the-art models, such as BERT and GPT-1, and enables us to utilize the large SWAG dataset for pre-training. The full dataset is available at \url{https://github.com/Websail-NU/CODAH}.
    \begin{table*}[ht]
    \centering\small
    \begin{tabular}{|c|c|c|}
    \hline
        \textbf{Category} & \textbf{Description} & \textbf{Example}\\
        \hline
        \hline
        Idioms&\begin{tabular}{@{}c@{}}Including phrases whose \\ meaning cannot be readily \\ interpreted from the meaning \\ of constituent parts\end{tabular}&\begin{tabular}{p{0.54\textwidth}}
        \textbf{A man on his first date wanted to break the ice. He} \\
        \hspace{0.5em} drank all of his water. \\
        \hspace{0.5em} threw the ice at the wall.	\\
        \hspace{0.5em} looked at the menu. \\
        \textbf{\hspace{0.5em} made a corny joke.}
        \end{tabular}\\
        \hline
        Negation&\begin{tabular}{@{}c@{}}Including negators to dictate \\ the
        meaning of the sentence\end{tabular}&\begin{tabular}{p{0.54\textwidth}}
        \textbf{The man's rebuttal was clearly not nonsensical. The rebuttal} \\
        \hspace{0.5em} has nothing to do with sense. \\
        \textbf{\hspace{0.5em} had some reasons associated with it.}	\\
        \hspace{0.5em} did not make any sense. \\
        \hspace{0.5em} was funny.
        \end{tabular}\\
        \hline
        Polysemy&\begin{tabular}{@{}c@{}}Testing the understanding of \\ multiple
        meanings of a \\ single word
        \end{tabular}&
        \begin{tabular}{p{0.54\textwidth}}
        \bf An architect retrieves his compass. He \\
        \hspace{0.5em} computes the area of a circle \\
        \hspace{0.5em} explores the open sea	\\
        \textbf{\hspace{0.5em} draws building dimensions on a canvas} \\
        \hspace{0.5em} uses his compass to find the north cardinal direction
        \end{tabular}\\
        \hline
        Reference&\begin{tabular}{@{}c@{}}Requiring understanding of \\ reference to one of multiple \\ subjects\end{tabular}&
        \begin{tabular}{p{0.54\textwidth}}
        \bf Rose is walking the dog while Joseph cooks dinner. Rose \\
        \hspace{0.5em} is following a new recipe. \\
        \hspace{0.5em} \textbf{enjoys the fresh air.}	\\
        \hspace{0.5em} wags her tail with joy. \\
        \hspace{0.5em} cuts tomatoes for the soup.
        \end{tabular}\\
        \hline\begin{tabular}{@{}c@{}}Quantitative\\ Reasoning\end{tabular}
        &\begin{tabular}{@{}c@{}}Involving basic arithmetic \\ calculations or comparisons
        \end{tabular}&
        \begin{tabular}{p{0.54\textwidth}}
        \bf A woman is walking two dogs and carrying a cat on her way to her car. She \\
        \textbf{\hspace{0.5em} puts all three animals in the back seat before driving off.} \\
        \hspace{0.5em} puts all four animals in the back seat before driving off.	\\
        \hspace{0.5em} puts both animals in the back seat before driving off. \\
        \hspace{0.5em} puts all nine animals in the back seat before driving off.
        \end{tabular}\\
    \hline
    \end{tabular}
    \caption{\label{labels-table} Question categories, descriptions, and examples}
    \end{table*}
    
    \subsection{Question Production}
    \label{sec:prod}
    We collected questions via a Web-based system. Participants were asked to compose a complete question, including the prompt, subject, and the four candidate completions. They would then be presented with the response of a pre-trained BERT model to their question. The pre-trained model consisted of a BERT-base model fine-tuned on the SWAG training set for 3 epochs with a batch size of 8. This model achieved 80.68\% accuracy on the SWAG validation set. The ability to obtain real-time feedback about the model's answers allowed participants to explore areas of weakness and design challenging questions. All submitted questions were added to the dataset, whether they fooled the baseline model or not.
    
    Annotators were provided explicit incentives to produce questions that the model answered incorrectly. The vast majority of submissions were contributed by university computer science students, who were familiar with neural network question answering systems.  Students were rewarded with extra credit points for submitting valid questions that fooled the baseline model.  Further, students could earn an equal number of extra credit points for questions that fooled the model when evaluated in cross-validation, after fine-tuning on other submitted questions.  This protocol was designed to encourage the creation of challenging and valid commonsense questions that are also free from stylistic annotation artifacts or redundancy, which would reduce the difficulty of the questions after fine-tuning and reduce the returns on their submissions. A small portion of the dataset was submitted anonymously by other individuals.

    We received a total of 4,149 raw questions, which were read and cleaned by four annotators (the authors). During cleaning, the answer choice order was shuffled and model's output answer were hidden from the annotator. We removed submissions with multiple or no distinctive commonsense answers, spelling or grammatical errors, incorrect answers, as well as duplicate submissions. The remaining questions were judged natural and easily answerable from common sense with minimal ambiguity and dispute. The cleaning operation produced our current 2,801-question dataset.
    
    Our 2,801-question dataset contains submissions from 116 named participants. The median, mean and standard deviation of the number of valid questions submitted by named individuals are 20.00, 21.38, and 13.86. The most prolific contributor submitted 86 questions. Anonymous participants contributed 321 questions, which is 11\% of the final dataset.
    
\section{Experiments}
    We evaluate the dataset on state-of-the-art neural question answering systems built on the BERT and GPT-1 architecture and provide multiple baselines. The models and experiment setups are discussed below. We also analyze the questions to identify distinctive categories of commonsense reasoning that provide a finer-grained understanding of model performances. In addition, the ablation experiments on dataset size and the use of fine-tuning on SWAG data allow us to further understand the impact of the relatively small size of CODAH.
    
    \subsection{Question Categorization}
    One of our goals is to analyze how system and human performance varies across questions in CODAH that employ different types of common sense. Therefore, we identified a small number of unambiguous categories of common sense, such as questions involving quantitative reasoning or negation. These categories only apply to a portion of the questions in our dataset, but have the advantage of being unambiguous and in many cases predictive of low system performance. In earlier attempts to devise categories to cover {\em all} questions, similar to analysis performed for textual entailment \cite{lobue2011types},  we found the inter-annotator agreement on such complete categorizations to be substantially lower (at \textless 0.4), even after iterating on category definitions. 
    
    We manually inspected all questions in our dataset and annotated each with one or more category labels, representing all types of reasoning required to identify the correct answer and eliminate incorrect ones. The descriptions and examples of these categories are found in Table \ref{labels-table}. Four human annotators (the authors) categorized the questions, and we calculated a Feiss' Kappa score of 0.63 between the annotators over an additional 50 questions. Table \ref{class-table} shows the distribution of labels over the entire dataset.
    
    \begin{table}[ht]
        \begin{center}
        \small
            \begin{tabular}{|l|c|c|}
            \hline \bf Category & \bf Count & \bf Percentage \\ \hline\hline
            Idioms & 249 &  8.8 \\ \hline
            Reference & 133 & 4.8 \\ \hline
            Polysemy & 108 & 3.9 \\ \hline
            Negation & 116 &  4.1 \\ \hline
            Quantitative & 87 & 3.1 \\ \hline
            Other & 2108 & 75.3  \\ \hline\hline
            Total & 2801  & \\ \hline
            \end{tabular}
        \end{center}
        \caption{\label{class-table} Distribution of question categories.}
    \end{table}
    
    \subsection{Models}
    \subsubsection{BERT}
    We evaluate a pre-trained BERT-Large \cite{devlin2018bert} implemented in PyTorch on the CODAH dataset. This model consists of a 24-layer network, with 1,024 hidden units per layer, 16-heads and a total of 340M parameters. For fine-tuning, we choose hyperparameters as described in Section~\ref{sec:bert_grid_search}, but in all cases we use a learning rate warmup over the first 10\% of training and linear learning rate decay.
    
    \subsubsection{OpenAI GPT-1}
    We also evaluate a pre-trained GPT model implemented in PyTorch. As described in \newcite{radford2018improving}, this model consists of a 12-layer decoder transformer with 12 attention heads and 3,072-dimensional hidden states. Our fine-tuning configuration is the same as described in the original paper: a batch size of 32, learning rate of 6.25e-5, linear learning rate decay over 3 epochs (with warmup over 0.2\% of training), and $\lambda$ of 0.5 (where $\lambda$ is a tuning coefficient that balances language-modeling loss and multiple-choice loss).

    \subsection{Model Evaluation}
    \label{sec:model_eval}
    We evaluate the models on several different train and test configurations described below. The CODAH dataset is evaluated in 5-fold stratified cross-validation which balances the distribution of question categories in each fold. 
    \begin{itemize}
        \item \textbf{CODAH:} Cross-validation fine-tuning on the CODAH dataset. The CODAH 80\% experiment represents the standard cross-validation setting on the full dataset, training on 80\% of the data in each fold and evaluating on the remaining 20\%. The 60\%, 40\% and 20\% ablation experiments are trained on a smaller portion of the CODAH dataset for each fold, but are evaluated in on the same test set which consists of 20\% of the full dataset. The question categories are balanced in both training set and test set. This makes the results from the experiments more comparable with each other. Three trials are conducted for all settings; the mean and standard deviation of the model accuracy are reported in Table \ref{table:ablation_results}.
        \item \textbf{SWAG+CODAH:} Fine-tuned on SWAG first, then fine-tuned again in cross-validation on CODAH. Ablation experiments are conducted in the same way as in the CODAH-only setting above, with the same dataset splits for training. The mean and standard deviation of the three trials are reported in Table \ref{table:ablation_results}.
        \item \textbf{SWAG only:} Fine-tuned on SWAG and evaluated on CODAH. Only one trial is conducted.
        \item \textbf{Answer only:} Cross-validation fine-tuning on the full CODAH dataset with the questions left blank (in both training and testing). Only one trial is conducted.
    \end{itemize}

    \subsection{BERT Hyperparameter Search}
    \label{sec:bert_grid_search}
    The authors of BERT suggest a hyperparameter grid search to perform when training BERT on new datasets. We perform a slightly modified version of this grid search in 5-fold cross validation over the training set in each fold of the cross-validation experiments described in Section~\ref{sec:model_eval}. That is, the complete grid search is run ten times, once for each fold of the CODAH and SWAG+CODAH settings. For the answer-only setting, we use the optimal hyperparameters found for the CODAH-only setting. Also, when training the initial SWAG model we use the hyperparameters recommended in the BERT paper, namely a batch size of 16, learning rate of 2e-5, and 3 epochs.

    In our initial experiments, we found that a lower learning rate and more training epochs produced higher accuracy on CODAH, so we replaced the 5e-5 learning rate in the original grid search with 1e-5, and we added a 6-epoch setting. The final hyperparameter grid is as follows:
    \begin{itemize}
        \item \textbf{Batch size:} 16, 32
        \vspace{-0.75em}
        \item \textbf{Learning rate:} 1e-5, 2e-5, 3e-5
        \vspace{-0.75em}
        \item \textbf{Number of epochs:} 3, 4, 6
    \end{itemize}

    In addition, we observed that in rare cases BERT fails to train; that is, after several training epochs it has accuracy approximately equal to that of random guessing. To combat this, any time BERT fails to achieve 30\% or greater accuracy on the test set for a given fold, we run it again with a different random seed. If it fails five times, take the average of those five trials as the final accuracy.

    \subsection{Results}
    Results for the above configurations are shown in Table \ref{table:ablation_results}. As a baseline, we evaluate both models on the full SWAG training and validation sets, providing an accuracy of 84.2\% on BERT and 80.2\% on GPT. To adjust for the difference in size between our dataset and SWAG, we also train the models on a sample of 2,241 SWAG questions (the size of the training set in each of CODAH's cross-validation folds) and evaluate them on the full SWAG validation set. This produces an accuracy of 75.2\% for BERT (using the cross-validation grid search) and 63.6\% for GPT.
    
    \begin{table}[ht]
        \begin{center}
        \small
            \begin{tabular}{|l|c|c|c|c|}
    			\hline \bf Training set    & \bf BERT \%                     & \bf GPT-1 \%  \\\hline\hline
			    CODAH 80\%               & 67.5 \hspace{0.1em} (1.24)    & 62.3 \hspace{0.1em} (1.11) \\\hline
			    CODAH 60\%               & 64.8 \hspace{0.1em} (2.24)    & 58.6 \hspace{0.1em} (0.65) \\\hline
			    CODAH 40\%               & 62.3 \hspace{0.1em} (1.34)    & 54.4 \hspace{0.1em} (1.27) \\\hline
			    CODAH 20\%               & 51.6 \hspace{0.1em} (1.56)    & 45.3 \hspace{0.1em} (0.76) \\\hline
			    SWAG+CODAH 80\%          & 69.5 \hspace{0.1em} (0.34)    & 65.5 \hspace{0.1em} (0.24) \\\hline
			    SWAG+CODAH 60\%          & 68.6 \hspace{0.1em} (0.49)    & 63.7 \hspace{0.1em} (0.26) \\\hline
			    SWAG+CODAH 40\%          & 65.8 \hspace{0.1em} (0.73)    & 61.2 \hspace{0.1em} (0.30) \\\hline
			    SWAG+CODAH 20\%          & 63.2 \hspace{0.1em} (0.70)    & 56.9 \hspace{0.1em} (0.07) \\\hline
			    SWAG only                & 41.3                          & 38.1  \\\hline
			    CODAH (Answer only)      & 52.2 \hspace{0.1em} (1.34)    & 53.4 \hspace{0.1em} (1.14)  \\\hline
            \end{tabular}
        \end{center}
        \caption{\label{table:ablation_results} Accuracy of BERT and GPT on different training settings when tested on CODAH. Numbers in parentheses represent the standard deviation.}
    \end{table}
    
    \subsection{Human Evaluation}
    For each category, we measure the accuracy of the BERT and GPT models trained on SWAG+CODAH. We also measure human accuracy as a baseline. Human accuracy was calculated as the mean accuracy of three human annotators, covering 707 dataset questions in total. Human annotators answered 95.3\% of questions correctly, presenting a 7-fold reduction in error compared to the fine-turned BERT model. Inter-annotator agreement was computed over a set of 50 additional questions with a pairwise average Cohen-Kappa score of 0.89, which is interpreted as almost perfect agreement by some guidelines. Table \ref{table:class_breakdown} displays the accuracy of the human annotators and neural networks on each category.
        
    \begin{table}[ht]
        \begin{center}
        \small
            \begin{tabular}{|l|c|c|c|}
            \hline \bf Category & \bf Human \% & \bf BERT \% & \bf GPT-1 \% \\ \hline\hline
                   Quantitative & 97.6         & 58.2 (3.69) & 48.7 (1.76) \\ \hline
                   Polysemy     & 91.7         & 63.9 (1.60) & 56.2 (2.87) \\ \hline
                   Negation     & 100          & 65.2 (2.17) & 62.4 (2.63) \\ \hline
                   Idioms       & 97.5         & 71.4 (1.23) & 71.9 (1.75) \\ \hline
                   Reference    & 100          & 72.4 (0.87) & 70.7 (1.99) \\ \hline
                   Other        & 94.9         & 70.2 (0.28) & 65.9 (0.31) \\ \hline\hline 
                   Total        & 95.3         & 69.5 (0.34) & 65.5 (0.24) \\ \hline
            \end{tabular}
        \end{center}
        \caption{\label{table:class_breakdown} Class-wise and overall accuracy of human annotators and neural network models, sorted by BERT performance on the proposed categories. Numbers in parentheses represent the standard deviation.}
    \end{table}
    
\section{Discussion}
    Based on our experiments, we find that model performance on CODAH is substantially lower than those seen on SWAG, which has seen models achieve over 85\% accuracy. We observed a decrease of 14.7\% on both models between the accuracy on SWAG and the accuracy on our SWAG+CODAH setting. This is especially significant since human error on CODAH is 4.7\%---less than a third of the 15\% expert error on the SWAG dataset. This suggests that CODAH is challenging to our QA systems because of the difficult commonsense reasoning involved, and not because of ambiguity or intractability in the dataset.
    
    \subsection{Question Categories}
    The logic categories including Quantitative and Negation are especially difficult for our models, seeing some of the lowest accuracies from both models, in contrast to the 99.0\% weighted average human accuracy on these categories. Surprisingly, both models performed very well on the Idioms category, suggesting that our neural systems may be capable of learning idioms just like other semantic knowledge.  Further identification of additional distinctive and interesting categories that cover the entire dataset may prove very useful in directing our efforts towards aspects of our commonsense QA systems that require the most attention.
    
    \subsection{Annotation Artifacts}
    Annotation artifacts are known to exist in many datasets and may be exploited by supervised models to achieve inflated performances \cite{gururangan2018annotation}. In CODAH, we did not explicitly filter questions with artifacts or try to detect them. We instead incentivize the question authors, who have some knowledge of how the learners work, to avoid introducing noticeable artifacts in their submissions, as explained in Section \ref{sec:prod}. Our results show that artifacts do not provide sufficient signal for state-of-the-art neural models to come close to human-level accuracy on our data. 
    
    \subsection{Answer-Only Baseline}
    In the answer-only experiment (where questions are omitted during training and testing), we found that BERT achieves 52.2\% accuracy and GPT-1 achieves 53.4\% accuracy, both of which are approximately the equivalent of narrowing four random options down to two. By comparing this to the CODAH experiment setting, we can interpret these results as an indication of the extent to which the signal was in the answers. While this could be due to artifacts, such as the right answer commonly being of a certain length, we also observed that in many cases, distinguishing between reasonable and ridiculous answers (without seeing the premise) is a part of commonsense reasoning. For example, a commonsense reasoner would be able to rule out the choice ``picks up his phone and calls his mom to tell her he doesn't have his phone" without seeing the premise, as a contradiction is contained in the answer. Similarly, ``kicks a field goal, celebrates by transforming into a fish, and then quits football" is unlikely to be veracious regardless of the hidden subject. 
    
    \subsection{Dataset Size}
    Our experiments show that CODAH forms a challenging extension to the existing SWAG dataset. Even when we train a system to perform near human-level on SWAG, and then fine-tune on CODAH, the system still struggles to answer CODAH questions correctly. However, CODAH is also smaller than SWAG, raising the question of whether CODAH remains more difficult than SWAG when training set size is equalized.  When we restrict to a subset of SWAG of the same number of questions as CODAH, we find that SWAG and CODAH have comparable accuracy for GPT (63.6\% on reduced-size SWAG vs. 62.3\% for CODAH) but on BERT there is larger gap (75.2\% accuracy on reduced-size SWAG, vs. 67.5\% accuracy on CODAH).  This suggests that CODAH questions are somewhat more difficult than SWAG questions when data set size is equalized, but the performance gap is much smaller than comparing with full-size SWAG. In addition, our ablation experiments where we train on different sizes of CODAH suggest that BERT and GPT performance improve with more data on both the CODAH-only and SWAG+CODAH settings, but the rate of improvement slows down as data size increases. The slowdown in improvement is more significant for BERT than GPT.
    
    Our results suggest two recommendations for dataset construction which we hope to evaluate in future work. The first is, rather than using a single protocol to collect one monolithic dataset, the community may be able to obtain more challenging data by aggregating a variety of distinct, independently-gathered datasets that follow a similar format. For example, pre-training on SWAG and evaluating on CODAH forms a more challenging benchmark than training and testing on SWAG alone.  Secondly, if we wish to use our adversarial collection approach to grow CODAH to tens of thousands of examples, we should update our system as new data arrives, so that contributors are able to tune their questions to remain difficult for the strongest, most up-to-date version of the system. Under such a data collection scheme, we may need to increase the reward for fooling the model in cross-validation compared to that for fooling the current model (whereas, these two rewards were equal in CODAH), in order to disincentivize adversarial attacks that manipulate the current model to make it easy to fool on subsequent questions.
    
\section{Conclusion}
    We present CODAH, a commonsense question answering dataset that is adversarially-constructed by allowing humans to view feedback from a pre-trained model and use this information to design challenging commonsense questions. Our experimental results show that CODAH questions present a complementary extension of the SWAG dataset, testing additional modes of common sense. 

    We identify specific categories of commonsense questions to determine types of reasoning that are more challenging for existing models. In particular, we note that Quantitative questions have low accuracy for both BERT and GPT. A more detailed analysis into why models struggle to reason about numbers as well as development of more detailed categories of commonsense reasoning are items for future work.
    
\section*{Acknowledgments}
We thank the anonymous reviewers, Yiben Yang, and Chandra Bhagavatula for helpful comments and feedback. We also thank the students of Northwestern EECS 349 Fall 2018, whose creativity and insight made this work possible. This work was supported in part by NSF Grant IIS-1351029 and the Allen Institute for Artificial Intelligence.

\bibliography{naaclhlt2019}
\bibliographystyle{acl_natbib}

\end{document}